\documentclass{elsarticle}

\usepackage{lineno,amsmath}

\journal{Journal of Biomedical Informatics}

\usepackage{multirow}
\usepackage{amsmath}
\usepackage{amssymb}
\usepackage{booktabs}
\usepackage{graphicx}
\usepackage{color}
\usepackage{longtable}
\usepackage{siunitx}
\usepackage{xurl}
\usepackage{threeparttable}
\usepackage[referable]{threeparttablex}
\usepackage{etoolbox}

\robustify\tnote
\robustify\bfseries

\usepackage[format=plain,font=scriptsize,labelfont=bf]{caption}
\usepackage{hyperref}
\hypersetup{pdfauthor={Name}}

\newcounter{ex}

\biboptions{numbers,sort&compress}

\definecolor{andrejcolor}{rgb}{0.7,0,0.7}

\begin{document}

\begin{frontmatter}

\title{Drug Repurposing for COVID-19 via Knowledge Graph Completion}


\author[add1]{Rui Zhang\corref{equalcontribute}\corref{mycorrespondingauthor}}
\author[add2]{Dimitar Hristovski\corref{equalcontribute}}
\author[add1]{Dalton Schutte\corref{equalcontribute}}
\author[add2]{Andrej Kastrin\corref{equalcontribute}}
\author[add3]{Marcelo Fiszman}
\author[uiuc]{Halil Kilicoglu}

\cortext[mycorrespondingauthor]{Corresponding author}
\cortext[equalcontribute]{Authors contributed equally}

\address[add1]{Institute for Health Informatics and Department of Pharmaceutical Care \& Health Systems, University of Minnesota, Minnesota, USA}
\address[add2]{Institute for Biostatistics and Medical Informatics, Faculty of Medicine, University of Ljubljana, Ljubljana, Slovenia}
\address[add3]{NITES - Núcleo de Inovação e Tecnologia Em Saúde, Pontifical Catholic University of Rio de Janeiro, Brazil}
\address[uiuc]{School of Information Sciences, University of Illinois at Urbana-Champaign, Champaign, IL, USA}

\begin{abstract}
\textbf{Objective.} To discover candidate drugs to repurpose for COVID-19 using literature-derived knowledge and knowledge graph completion methods.\\
\textbf{Methods.} We propose a novel, integrative, and neural network-based literature-based discovery (LBD) approach to identify drug candidates from PubMed and other COVID-19-focused research literature. Our approach relies on semantic triples extracted using SemRep (via SemMedDB). We identified an informative and accurate subset of semantic triples using filtering rules and an accuracy classifier developed on a BERT variant. We used this subset to construct a knowledge graph, and applied five state-of-the-art, neural knowledge graph completion algorithms (TransE, RotatE, DistMult, ComplEx, and STELP) to predict drug repurposing candidates. The models were trained and assessed using a time slicing approach and the predicted drugs were compared with a list of drugs reported in the literature and evaluated in clinical trials. These models were complemented by a discovery pattern-based approach.\\
\textbf{Results.} Accuracy classifier based on PubMedBERT achieved the best performance (F$_1$ = 0.854) in classifying semantic predications. Among five knowledge graph completion models, TransE outperformed others (MR = 0.923, Hits@1 = 0.417). Some known drugs linked to COVID-19 in the literature were identified, as well as others that have not yet been studied. \textcolor{black}{Discovery patterns enabled identification of additional candidate drugs and generation of plausible hypotheses regarding the links between the candidate drugs and COVID-19. Among them, five highly ranked and novel drugs (paclitaxel, SB 203580, alpha 2-antiplasmin, metoclopramide, and oxymatrine) and the mechanistic explanations for their potential use are further discussed.}\\
\textbf{Conclusion.} \textcolor{black}{We showed that a LBD approach can be feasible not only for discovering drug candidates for COVID-19, but also for generating mechanistic explanations.} Our approach can be generalized to other diseases as well as to other clinical questions. \textcolor{black}{Source code and data are available at \url{https://github.com/kilicogluh/lbd-covid}.}
\end{abstract}



\end{frontmatter}

\section{Introduction}
Coronavirus disease 2019 (COVID-19), caused by a novel coronavirus named severe acute respiratory syndrome coronavirus 2 (SARS-CoV-2; formerly 2019-nCoV), first emerged in China in late 2019, and was declared a global pandemic by the World Health Organization (WHO) on March 11, 2020. Since then, COVID-19 has disrupted human life across the globe, with enormous human, economic, and societal costs. \textcolor{black}{At the time of writing, it shows no sign of abating \cite{whocovid,jhucovid}, although the final months of 2020 have brought some good news. First, on October 22, 2020, after the initial submission of this manuscript, the Food and Drug Administration (FDA) approved remdesivir for the treatment of COVID-19 requiring hospitalization~\cite{remdesivir_fda}. Then, on November 9, 2020, Pfizer/BioNTech announced the effectiveness of their coronavirus vaccine BNT162b2 and over a month later after the release of additional data, FDA granted it emergency use authorization \cite{vaccine_fda}. A second vaccine, by Moderna, has also been authorized for emergency use on December 18, 2020 \cite{vaccine_fda_moderna}.}


Rapid development of effective vaccines for COVID-19 was by no means guaranteed, however. Moreover, \emph{de novo} development and approval of an effective antiviral therapy remains a risky, costly, and time-consuming process. In the absence of an effective vaccine or other therapies, there have been significant efforts in repurposing drugs approved for other diseases for COVID-19 treatment, some of which have been tested in clinical trials (e.g., dexamethasone \cite{recovery20}, hydroxychloroquine and lopinavir/ritonavir \cite{horby20}) \textcolor{black}{and one ultimately approved by FDA for treatment of patients hospitalized with COVID-19 (remdesivir \cite{remdesivir_fda,beigel20}).}


Computational approaches to drug repurposing have also garnered much attention to accelerate discovery of therapies for COVID-19 \cite{altay20,wang20covid}. Common computational drug repurposing methods include drug signature matching, molecular docking, genome-wide association studies, and network analysis \cite{pushpakom19}. These data-driven approaches involve systematic analysis of various types of biological and clinical data (e.g., gene expression, chemical structure, genome and protein sequences, and electronic health records) to generate hypotheses regarding repurposed use of approved or investigational drugs \cite{pushpakom19}. The potential of recent advances in artificial intelligence (AI) and machine learning for COVID-19 drug repurposing has also been highlighted \cite{zhou20ai} and several studies using these techniques have reported promising results \cite{ge20,zhou20network,zhou20networkb,zeng20}. In particular, approaches leveraging network medicine \cite{barabasi11} principles and biological knowledge graphs have been emphasized \cite{zhou20ai}.

Most of the computational approaches to drug repurposing have focused on biological data, such as gene expression,  protein-protein and drug-target interactions, and used SARS-CoV-2-related data. However, COVID-19-specific data is meaningful in the context of the larger body of diverse knowledge underpinning medicine and life sciences, a primary source of which is the biomedical literature. While some COVID-19 drug repurposing studies incorporated literature-based knowledge~\cite{ge20,zeng20}, their focus has remained largely COVID-19-specific. We argue that efficiently and safely repurposing drugs to treat COVID-19 requires more effective integration of literature-based knowledge with biological data collected via high-throughput methods.

In this paper, we propose a novel literature-based discovery \cite{henry17,sebastian17} approach for COVID-19 drug repurposing. Similar to related work \cite{zeng20}, we cast drug repurposing as a task of knowledge graph completion (or link prediction). We use a large, literature-derived biomedical knowledge graph constructed from SemMedDB \cite{semmeddb} as well as COVID-19 research literature \cite{cord19}, as our data source. We use several state-of-the-art, neural network-based algorithms \cite{transe,rotate,distmult,complex,stelp} for the task, and also complement these approaches with an approach based on discovery patterns \cite{hristovski06}. \textcolor{black}{Furthermore, we highlight the role of discovery patterns in search of mechanistic explanations for candidate drugs.} Unlike most approaches that focus on COVID-19-specific knowledge \cite{ge20,zeng20}, we consider a larger body of biomedical knowledge, as captured in the PubMed bibliographic database as well as in the COVID-19 research literature. Our results show that our approach can identify known drugs that have been used for COVID-19 and discover other novel drugs that can potentially be repurposed for COVID-19.

\section{Related Work}
\subsection{COVID-19 computational drug repurposing} 
Significant computational work has already been done to prioritize FDA-approved drugs for repurposing to treat COVID-19 \cite{altay20,wang20covid}. For the most part, these studies can be categorized as molecular docking-based drug screening studies and network-based studies, the majority of them belonging to the former category. In molecular docking studies, small molecules in compound libraries are screened for effectiveness against the host proteins in the SARS-CoV-2 host interactome. Many studies of this kind have been reported, and some of the proposed drugs such as ritonavir, ribavirin, remdesivir, and oseltamivir have been used in practice and many are being evaluated in clinical trials \cite{gordon20,riva20,wu20analysis,elfiky20,kandeel20,alkhafaji20,wang20fast,elfiky20b}.

While not as common as docking studies, network-based approaches to drug repurposing have also been explored. In one early study, a virus-related knowledge graph which consists of drug-target and protein-protein interactions and similarity networks from publicly available databases (e.g., DrugBank~\cite{drugbank}, ChEMBL~\cite{chembl}, BioGRID~\cite{biogrid}) was constructed and network-based machine learning and statistical analysis were used to predict an initial list of COVID-19 drug candidates. This list was narrowed down based on text mining from the literature and gene expression profiles from COVID-19 patients, and a poly-ADP-ribose polymerase 1 (PARP1) inhibitor CVL218,  was proposed for therapeutic use against COVID-19 ~\cite{ge20}. Cava et al.~\cite{cava20} used gene expression profiles from public datasets to construct a protein-protein interaction network in conjunction with pathway enrichment analysis to identify 36 potential drugs, including nimesulide, thiabendazole, and fluticasone propionate. In another study, network proximity analyses of drug targets and HCoV-host interactions in the human interactome were used to prioritize 16 potential repurposed drugs, including melatonin, mercaptopurine, and sirolimus, which were validated by enrichment analyses of drug-gene signatures and transcriptome data in human cell lines~\cite{zhou20network}. Potentially useful drug combinations (e.g., melatonin plus mercaptopurine) were also suggested. A follow-up study combined network medicine approaches based on human interactome with clinical patient data from a COVID-19 registry to show that melatonin was associated with reduced likelihood of a positive SARS-CoV-2 laboratory test~\cite{zhou20networkb}. The approach was further extended to explore deep learning~\cite{zeng20}. A comprehensive knowledge graph of drugs, diseases, and proteins/genes (named CoV-KGE) was constructed by combining molecular interaction information from the literature with knowledge from DrugBank. A knowledge graph embedding model, named RotatE~\cite{rotate} was used to represent the entities and the relationships in the knowledge-based in low-dimensional vector space. Using the ongoing COVID-19 trial data as a validation set, 41 high-confidence repurposed drug candidates (including dexamethasone, indomethacin, niclosamine, and toremifene) were identified, and further validated via an enrichment analysis of gene expression and proteomics data in SARS-CoV-2-infected human cells. Another study used node2vec graph embeddings and variational graph autoencoders for the same purpose~\cite{ray20covid}. Gysi et al.~\cite{gysi20} evaluated three algorithms (graph neural network, network proximity, and network diffusion) on a network of drug protein targets and disease-associated proteins for COVID-19 drug repurposing. While they obtained low correlations across the three algorithms, an ensembling approach that combined the predictions of all algorithms was shown to outperform the individual methods, ranking ritonavir, chloroquine, and dexamethasone among the most promising candidates. Some limited literature knowledge relevant to COVID-19 has been incorporated to network-based approaches; however, their focus remains largely on structured molecular interaction information encoded in databases.

\subsection{Literature-based discovery}

Literature-based discovery (LBD) \cite{henry17,sebastian17} is a method of automatic hypothesis generation pioneered by Swanson~\cite{swanson86}. Based on the concept of ``undiscovered public knowledge'', LBD seeks to uncover valuable hidden connections between disparate research literatures, and has been proposed as a potential solution for the problem of ``research silos'' (the view that scientific research areas are largely isolated from one another). The primary LBD paradigm is the so-called ABC model. In the \emph{open discovery} form of this model, a relationship between two concepts A and B is known in one research area and another relationship between concepts B and C is known in another, and a potential relationship between concepts A and C is proposed. Conversely, in \emph{closed discovery}, relationship AC is known, and a concept B is proposed as an explanation for the relationship AC. Extensions to ABC model have also been proposed, such as discovery browsing model that aims to elucidate more complex relationship paths between biomedical concepts~\cite{wilkowski11,cairelli13}. Most applications of LBD have been in the biomedical domain, beginning with Swanson’s discovery of fish oil as a treatment for Raynaud disease \cite{swanson86}, a hypothesis supported subsequently by clinical studies. While early LBD systems focused primarily on term co-occurrence~\cite{swanson97,weeber01}, semantic relations have been widely used in later years for representing scientific content of biomedical publications~\cite{hristovski06,ahlers07lbd,preiss15, cameron15}. More recently, distributed vector representations based on term or semantic relation co-occurrence have been gaining popularity~\cite{cohen10rri,cohen11sim,cohen17}.

Drug repurposing has been one of the prominent applications of LBD~\cite{hristovski06,hristovski10,hristovski13,cohen14,zhang14b,rastegar15,yang17lbd}. For example, Hristovski et al.~\cite{hristovski06} used semantic discovery patterns following the ABC model to identify potential therapeutic uses for drugs. Zhang et al.~\cite{zhang14b} used discovery patterns and SemMedDB relations to identify potential prostate cancer drugs. Cohen et al.~\cite{cohen14} used a vector representation approach based on semantic relations to predict a small number of active agents within a large library screened for activity against prostate cancer cells. 

\subsection{Knowledge graph completion}
Knowledge graphs are represented as a collection of head entity-relation-tail entity triples $(h,r,t)$, where entities correspond to nodes and relations to edges between them. Knowledge graph completion is the task of predicting unseen relations between two existing entities or to predict the tail entity given the head entity and the relation (or head entity given the tail entity and the relation). Recent approaches to knowledge graph completion rely on knowledge graph embedding methods~\cite{wang17kg}, which learn a mapping from nodes and edges to continuous vector space that preserve the proximity structure of the knowledge graph and are amenable to application of machine learning methods. Such methods include translational models, which use distance-based scoring functions (e.g., TransE~\cite{transe}, TransH~\cite{transh}, RotatE~\cite{rotate}), and semantic matching models, which use similarity-based scoring functions (e.g., RESCAL~\cite{rescal} , DistMult~\cite{distmult}, ComplEx~\cite{complex}, and HolE~\cite{nickel16hole}). Graph convolutional networks~\cite{dettmers17,schlichtkrull18} as well as methods that use context-based encoding approach (KG-BERT \cite{kgbert}, STELP \cite{stelp}) have also been recently proposed. Knowledge graph embedding techniques based on a network of drug, disease, and gene/protein entities have been used to support drug repurposing for rare diseases~\cite{sosa20}. 
Graph convolutional networks were used to model drug side effects resulting from drug-drug interactions~\cite{zitnik18}. A multimodal graph of protein-protein, drug-protein target, and drug-drug interactions was constructed from publicly available datasets. Sang et al.~\cite{sang18} constructed low-dimensional knowledge graph embeddings from SemMedDB relations and trained a Long Short-Term Memory (LSTM) model using known drug therapies from Therapeutic Target Database~\cite{chen02ttd} to propose potential drugs using the trained model.

\section{Materials and Methods}
\begin{figure}[htb!]
\fbox{\includegraphics[width=\textwidth]{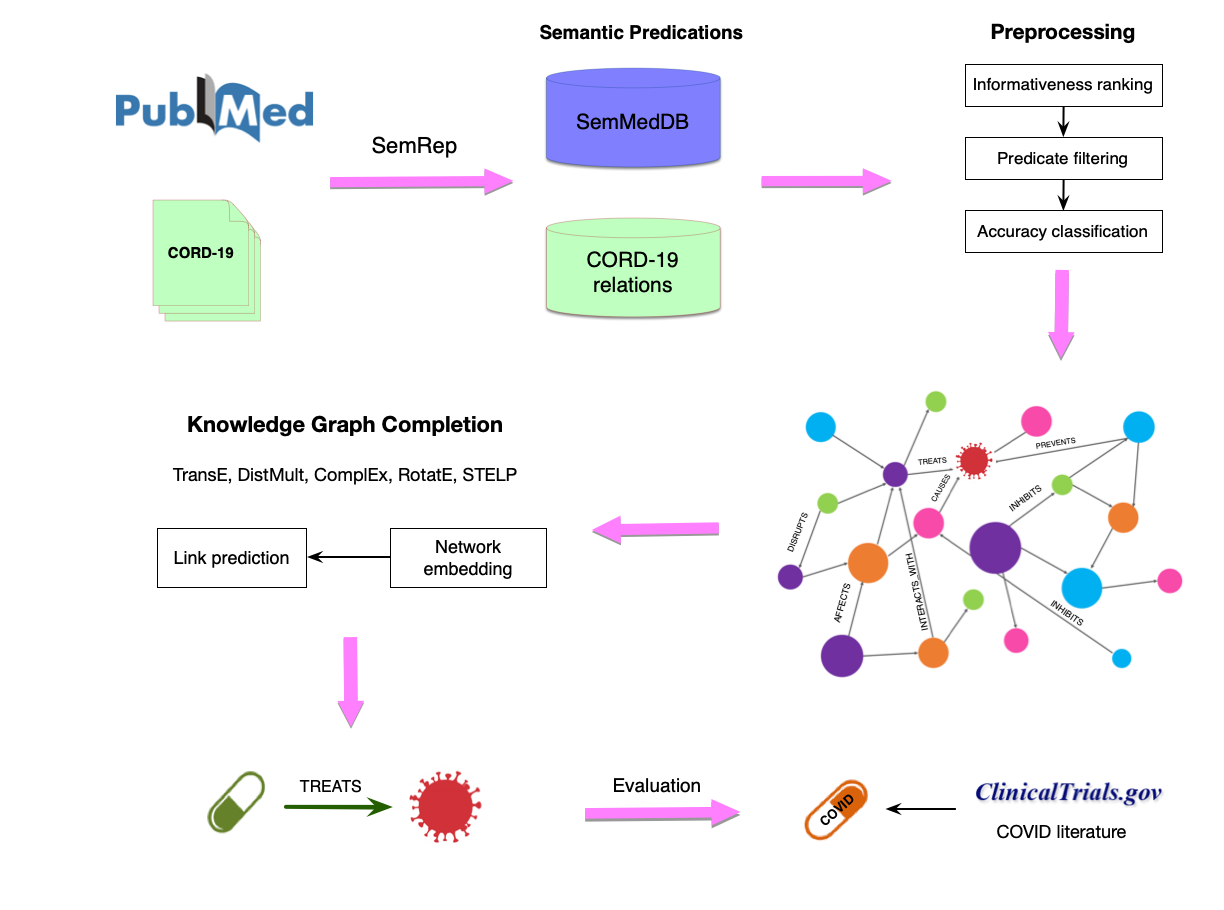}}
  \caption{Diagram illustrating the workflow of our approach.}
\label{fig1}
\end{figure}

In this section, we first describe our data sources and the preprocessing steps that were taken to construct a literature knowledge graph from these data sources. Next, we discuss the knowledge graph completion methods that we used to predict candidate drugs for COVID-19 as well as the discovery patterns used for providing mechanistic explanations. Lastly, we detail various evaluation schemes that we used to validate our predictions. A workflow diagram illustrating our approach is provided in Fig.~\ref{fig1}. Our source code and data are publicly available at \url{https://github.com/kilicogluh/lbd-covid}.

\subsection{Data}
We constructed our biomedical knowledge graph primarily from SemMedDB \cite{semmeddb}, a repository of semantic relations automatically extracted from biomedical literature using SemRep natural language processing (NLP) tool~\cite{semrep03,kilicoglu20semrep}. SemRep-extracted relations are in the form of subject-predicate-object triples (also called \emph{semantic predications}) and are derived from unstructured text in PubMed citations (i.e., titles and abstracts). For example, the triples \texttt{chloro\-quine}-\textsc{treats}-\texttt{Malaria} and \texttt{hydroxychloroquine}-\textsc{treats}-\texttt{Malaria} are extract\-ed from the fragment \emph{Chloroquine (CQ) and Hydroxychloroquine (HCQ) have been commonly used for the treatment and prevention of malaria} (PMID: 3291\-0933). Subject and object arguments are normalized to concept unique identifiers (CUIs) in the UMLS (Unified Medical Language System) Metathesaurus~\cite{umls,bodenreider04}. 
Concepts are enriched with UMLS semantic type information (Disease or Syndrome, Pharmacologic Substance, etc.) and the relations are linked to the supporting article and sentence. SemMedDB has supported a wide range of computational applications, ranging from gene regulatory network inference~\cite{chen14} to \emph{in silico} screening for drug repurposing~\cite{cohen14} and medical reasoning~\cite{origami}, and has also found widespread use for literature-based knowledge discovery and hypothesis generation (e.g.,~\cite{cairelli13,kastrin14,preiss15,sybrandt18,rindflesch18}).  In its most recent release (version 43, dated 8/28/2020)\footnote{\url{https://ii.nlm.nih.gov/SemRep\_SemMedDB\_SKR/SemMedDB/SemMedDB\_download.shtml}}, SemMedDB contains more than 107M relations from more than 31M PubMed citations and 209M sentences. This release also includes COVID-19-related concepts and, thus, can serve as a knowledge graph for COVID-19 drug repurposing.

COVID-19 literature has grown at an unprecedented rate. LitCovid, NCBI’s bibliographic database for COVID-19 literature~\cite{litcovid} contains over \textcolor{black}{82K articles (as of 12/21/2020)}. An even richer dataset is the COVID-19 Open Research Dataset (CORD-19), which contains over 200K articles (including historical research on coronaviruses)~\cite{cord19}. Not all of these articles are included in PubMed. To ensure that our knowledge graph provides adequate coverage of COVID-19 knowledge, we included CORD-19 articles not included in PubMed, as well, and used SemRep to extract relations from titles and abstracts of these articles. We used CORD-19 release dated 09/25/2020. 

SemMedDB distribution contains \num{107645218} relations among \num{339638} concepts. CORD-19 dataset processed through SemRep contains \num{505968} relations among \num{41609} concepts. 


\subsection{Preprocessing}
In this work, we focused on a subset of semantic relations derived from the combination of PubMed and CORD-19 datasets, predicted to be accurate and informative for drug repurposing. 

First, we eliminated relations involving generic biomedical concepts (i.e., relations in which both subject and object were present in the \texttt{GENERIC\_CONCEPT} table of SemMedDB such as \texttt{Pharmaceutical Preparations}) and relations with identical subject and object arguments. Next, we excluded a subset of predicate types that were not expected to be useful for drug repurposing, such as \textsc{part\_of} and \textsc{process\_of}. The predicate types we used are \textsc{affects}, \textsc{associated\_with}, \textsc{augments}, \textsc{causes}, \textsc{coexists\_with}, \textsc{complicates}, \textsc{disrupts}, \textsc{inhibits}, \textsc{interacts\_with}, \textsc{manifestation\_of}, \textsc{pre\-disposes}, \textsc{prevents}, \textsc{produces}, \textsc{stimulates}, and \textsc{treats}. Lastly, we also excluded the relations in which the subject or the object belongs to one of the following semantic groups: Activities \& Behaviors, Concepts \& Ideas, Objects, Occupations, Organizations, and Phenomena. The combined knowledge graph (SemMedDB + CORD-19) consists of \num{331427} unique nodes and \num{20017236} relations.

\textcolor{black}{In the second step, we eliminated (i) high-degree concepts using network degree centrality and (ii) uninformative semantic relations using log-likelihood ratio. The adjacency matrix $A$ of a knowledge graph (i.e., directed network with multiedges) with $n$ nodes (i.e., concepts) has entries $A_{ij} = 1$ if there is a relation from concept $i$ to concept $j$. The in- and out-degrees of concept $i$ can then be expressed as~\cite{boccaletti2006complex}:
\begin{equation*}
   k_{i}^{\text{in}} = \sum_{j = 1}^{n} A_{ji}
    \quad
    \text{and}
    \quad
     k_{i}^{\text{out}} = \sum_{j = 1}^{n} A_{ij}.
\end{equation*}
}
\textcolor{black}{To filter out uninformative links, we assigned} each semantic relation a $G^2$ score indicating how strongly the terms within a triple are associated with each other~\cite{McInnes2004}. A high $G^2$ score means that the observed and expected frequencies are significantly different, indicating that the triple is less likely to occur by chance. For computational purposes, we created two three-dimensional contingency tables with indices $i$, $j$, and $k$. The first table (OT) holds observed frequencies of a triple from the knowledge graph and the second table (ET) contains the expected values assuming independence of terms in each triple. $G^2$ was then calculated using the equation
\begin{equation*}
\label{eq:g2}
G^2 = 2 \times \sum_{i,j,k} n_{ijk} \times \log \left( \frac{n_{ijk}}{m_{ijk}} \right), \quad m_{ijk} = \frac{\sum_{i} n_{jk} \times \sum_{j} n_{ik} \times \sum_{k} n_{ij}}{T^2},
\end{equation*}
where $n_{ijk}$ is the cell $i,j,k$ in OT, $m_{ijk}$ is the cell $i,j,k$ in ET, and $T = \sum n_{ijk}$.

Next, we normalized all three measures ($G^2$, $k_{i}^{\text{in}}$, and $k_{i}^{\text{out}}$) to the range $[0, 1]$ and summed them up into a final score. The lower the score, the more specific and informative the relation is. For example, the relation \texttt{Operative Surgical Procedures}-\textsc{treats}-\texttt{Woman} which has a high score is more general than relation \texttt{interleukin-6}-\textsc{affects}-\texttt{Autoimmune Diseases}. We also kept all relations with biomedical concepts that refer to COVID-19 terms in the UMLS\footnote{\url{https://metamap.nlm.nih.gov/Covid19Terms.shtml}}:
\textcolor{black}{
\begin{itemize}
    \item \texttt{C5203670:COVID19 (disease)}
    \item \texttt{C5203671:Suspected COVID-19}
    \item \texttt{C5203672:SARS-CoV-2 vaccination}
    \item \texttt{C5203673:Antigen of SARS-CoV-2}
    \item \texttt{C5203674:Antibody to SARS-CoV-2}
    \item \texttt{C5203675:Exposure to SARS-CoV-2}
    \item \texttt{C5203676:SARS-CoV-2}
\end{itemize}
}
\textcolor{black}{We estimated that approximately 2.5M relations could be processed in reasonable amount of time with out GPU and eliminated relations with high final scores.}
At the end of the preprocessing stage, the knowledge graph consists of \num{131355} nodes and \num{2558935} relations. 

\subsubsection{Accuracy Classification}
The precision of semantic predications generated by SemRep vary by domain (e.g., clinical relationships are more precise than molecular interactions). To improve the precision of the relations used for drug repurposing, we extended the SemRep accuracy classifiers previously proposed \cite{zhang15mining,vasilakes18}. We fine-tuned a collection of Transformer-based pretrained language models to classify semantic predications as correct vs. incorrect. We used the following models: vanilla BERT (base size, cased and uncased)~\cite{bert}, BioBERT~\cite{biobert}, BioClinicalBERT~\cite{alsentzer19}, BlueBERT~\cite{bluebert}, and PubMedBERT~\cite{pubmedbert}.

To extend the coverage of our existing classifiers, we used \num{6492} predications annotated as correct vs. incorrect with respect to their source sentences. We leveraged \num{6000} annotations from a previous study \cite{vasilakes18} (Cohen's $\kappa = 0.80$) and annotated 492 new predications. Annotation guidelines generated in the previous study was used. 
Two of the authors (HK and MF) and two health informatics graduate students annotated predications containing predicates of interest absent in the prior study (Fleiss' $\kappa = 0.41$, indicating moderate agreement). \textcolor{black}{Fleiss' $\kappa$ was used in this case, as more than two annotators were involved in annotation~\cite{fleiss1971measuring}.} 

The resulting annotated set was split into 80/10/10 training/validation/test sets. Hyperparameters were determined empirically and the learning rate was set to $1 \times 10^{-5}$, the batch size was 16, the maximum number of epochs was set to 10 but early stopping was employed. Optimization was done using the Adam optimizer~\cite{adam} with decoupled weight decay regularization using betas (0.9, 0.999) and decay 0.01. The pooled output from the BERT model was fed through a linear layer to produce logits that then underwent a softmax transformation to return class probabilities. A single Tesla V100 GPU was used to train the models. We compared the performance of various above-mentioned transformers. The best classifier was then used to filter incorrect semantic predications.
\textcolor{black}{This resulted in \num{1016124} relations being kept for the knowledge graph completion methods.}

\subsection{Knowledge Graph Completion}
Consider a knowledge graph $\mathcal{G} = (\mathcal{E}, \mathcal{R}, \mathcal{E})$, where $\mathcal{E}$ refers to a set of entities, $\mathcal{R}$ denotes a set of possible relations, and $\mathcal{T}$ stands for a set of triples in the form (h)ead-(r)elation-(t)ail, formally denoted as $\{ (h, r, t) \} \subset \mathcal{E} \times \mathcal{R} \times \mathcal{E}$. The aim of knowledge graph completion is to infer new triples $(h^{\prime}, r^{\prime}, t^{\prime})$ such that $h^{\prime}, t^{\prime} \in \mathcal{E}$ and $r^{\prime} \in \mathcal{R}$. In this setting, the knowledge graph completion problem could be represented as a ranking task in which a prediction function $\psi (h, r, t) \colon \mathcal{E} \times \mathcal{R} \times \mathcal{E} \mapsto \mathbb{R}$ which generates higher scores for true triples and lower scores for false triples is learned.

We explored three classes of knowledge graph completion methods:  TransE \cite{transe} and RotatE~\cite{rotate} for translational models, DistMult~\cite{distmult} and ComplEx~\cite{complex} for semantic matching models, and STELP~\cite{stelp} for context-based encoding. These methods differ in the way that they encode entities and relations in a knowledge graph into a low-dimensional vector space (i.e., knowledge graph embedding). Such distributed vector representations can be used for downstream reasoning and machine learning tasks.

\subsubsection{Translational models (TransE and RotatE)}

\begin{figure}[htb!]
\centering
\includegraphics[scale=.7]{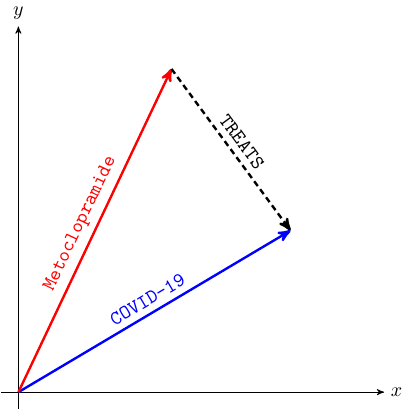}
\caption{\textcolor{black}{TransE models relations as translations on a low-dimensional embedding of the entities. If $(h,r,t)$ is true, the embedding of the tail entity $t$ (i.e., \texttt{COVID-19}) should be close to the embedding of the head entity $h$ (i.e., \texttt{Metoclopramide}) plus vector that depends on the relationship $r$ (i.e., \textsc{TREATS}).}}
\label{fig:fig2}
\end{figure}

TransE~\cite{transe} describes a triplet $(h, r, t)$ as a translation between head entity $h$ and tail entity $t$ through relation $r$ in a continuous vector space, i.e., $\mathbf{h} + \mathbf{r} \approx \mathbf{t}$, where $\mathbf{h}, \mathbf{r}, \mathbf{t} \in \mathbb{R}^d$ is the embedding of $h$, $r$, and $t$, respectively. To measure plausibility of relations TransE employs a distance-based score function $s(h,r,t) = \| \mathbf{h} + \mathbf{r} - \mathbf{t} \|$. Either $L_1$ or $L_2$ norm can be employed. \textcolor{black}{Figure~\ref{fig:fig2} illustrates TransE model with two-dimensional embedding.}

We choose TransE because of its simplicity and good prediction performance. However, TransE is able to model only one-to-one relations and fails to embed one-to-many, many-to-one, and many-to-many relations. To solve this problem, several other solutions have been proposed including RotatE~\cite{rotate}. RotatE treats each relation in a complex vector space as a rotation from the head entity to the tail entity, i.e., $s(h,r,t) = \vert \mathbf{h} \circ \mathbf{r} - \mathbf{t} \vert_{l_1}$, where $\circ$ is a Hadamard product. We selected  RotatE as a counterpart to TransE, as TransE reportedly does not perform well on some data sets (e.g., FB15k benchmark data set~\cite{transe}, commonly used in knowledge graph completion), which require symmetric pattern modeling.

\subsubsection{Semantic matching models (DistMult and ComplEx)}
DistMult~\cite{distmult} is the simplest approach among semantic matching models. Its scoring function is defined as $s(h,r,t) = \langle \mathbf{h}, \mathbf{r}, \mathbf{t} \rangle$. However, DistMult is limited only to symmetric relations, generating same scores for triples $(h,r,t)$ and $(t,r,h)$. ComplEx~\cite{complex} extends DistMult to the complex domain. Head and tail embeddings for the same entity are complex conjugates, enabling ComplEx to model asymmetric relations. Its score function is defined as $s(h,r,t) = \mathrm{Re} (\langle \mathbf{h}, \mathbf{r}, \bar{\mathbf{t}} \rangle)$, where $\mathbf{h}, \mathbf{r}, \mathbf{t} \in \mathbb{C}^k$, $\mathrm{Re}(\cdot)$ is a real part of a complex vector, and $k$ is dimension of an embedding.

Hyperparameters for both sets of models were tuned using the grid search on the validation set for each prediction model. We tuned the learning rate $\eta \in \{0.001, 0.01, 0.1\}$, number of hidden dimensions $k \in \{50, 100, 250, 400\}$, regularization coefficient $\lambda \in \{2 \times 10^{-6}, 2 \times 10^{-8}\}$, negative adversarial sampling $\in \{\mathtt{True}, \mathtt{False}\}$, fixed margin $\gamma \in \{1, 5, 10, 20\}$ for RotatE and norm $d \in \{ L_1, L_2 \}$ for TransE model. 

\subsubsection{Context-encoding models (STELP)}
Semantic Triple Encoder for Link Prediction (STELP)~\cite{stelp}, is a context-based encoding approach to knowledge graph completion. At its core is a Siamese BERT model that leverages sharing one set of weights across two models to produce encoded, contextual representations of the relations that are then fed to either a multi-layer perceptron (MLP) for classification or a similarity function for contrasting. The STELP architecture uses two learning objectives for training: triple classification and triple contrasting. The final learning objective is a linear combination of the two. During training, STELP takes a single positive relation $(h, r, t)$, and produces five negative relations $(h, r, t')$ by corrupting the tail. The head context, $(h, r)$, term is sent into one BERT model while each tail, $(t)$ or $(t')$, are each sent to the other BERT model that shares weights with the other. The classification objective seeks to classify $(h, r, t)$  as 1 and each $(h, r, t')$ as 0 while the contrastive objective seeks to measure the distance between the contextual embedding of the head and tail portions in a learned semantic space. Formally, the classification loss and constrastive loss functions are as follows:
\begin{align*}
    \mathcal{L}^c &= \frac{-1}{|\mathcal{D}|}\sum_{tp\in\mathcal{D}}\frac{1}{1+|\mathcal{N}(tp)|}\left(\log{s^c}+\sum_{tp'\in\mathcal{N}(tp)}\log{(1-s^{c'})} \right)\\
    \mathcal{L}^d &= \frac{1}{|\mathcal{D}|}\sum_{tp\in\mathcal{D}}\frac{1}{|\mathcal{N}(tp)|}\sum_{tp'\in\mathcal{N}(tp)}\max({0,\lambda-s^d+s^{d'}})
\end{align*}
where $\mathcal{D}$ is the set of correct triples, $\mathcal{N}(tp)$ is the set of corrupted triples for given positive triple $tp$, $s^c$ and $(1-s^{c'})$ are the positive class probability for $tp$ and negative class probability for $tp'$, respectively, $\lambda$ is the margin size, $s^d$ and $s^{d'}$ are the negative Euclidean distances between the contextual embeddings for the head and tail portions of the triple. The complete multi-objective loss function then is:
\begin{equation*}
\mathcal{L} = \mathcal{L}^c + \gamma \mathcal{L}^d
\end{equation*}
where $\gamma$ is a scaling factor for the contribution of the contrastive loss.

At inference, STELP considers every entity-context combination for a given partial relation, $(h, r)$ to find $(t)$ or $(r, t)$ to find $(h)$, and ranks every pair using the sum of the positive class probability and the scaled negative Euclidean distance.

\begin{figure}
\centering
\includegraphics[width=\textwidth]{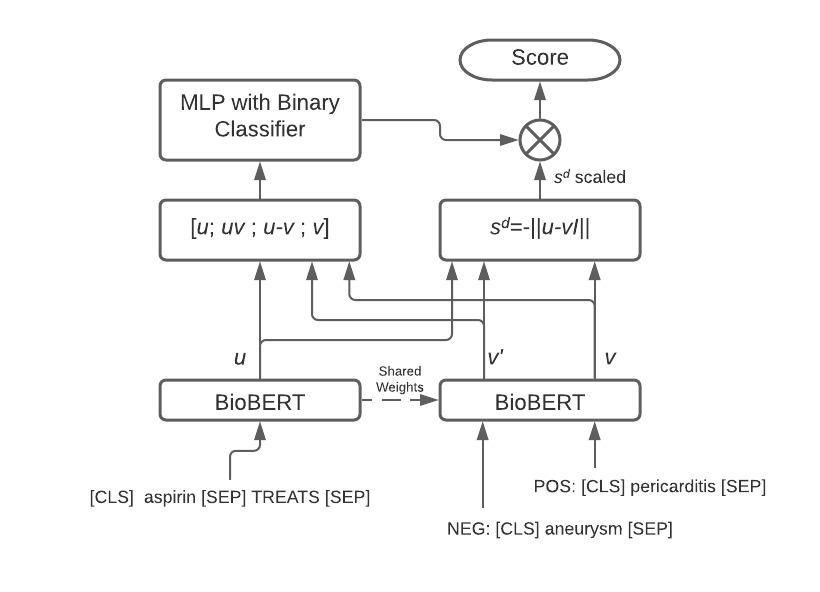}
\caption{\textcolor{black}{Diagram for the high-level architecture of STELP.}}
\label{fig:fig3}
\end{figure}

We replaced the vanilla base BERT model proposed in the STELP paper with BioBERT, trained on biomedical literature corpora.  The \num{1016124} unique relations remaining after preprocessing were each corrupted to produce five negative relations for a total of \num{5080620} negative relations and a grand total of \num{6096744} relations. The hyperparameters were set to the same values as in the original STELP paper and the learning rate was set to $1 \times 10^{-5}$, the batch size was 16, the contrastive loss scaling factor was 1.0. Optimization was done using Adam with decoupled weight decay with betas (0.9, 0.999) and decay 0.01. Training was run for \textcolor{black}{\num{190523}} training iterations. Ranking was done by adding the scaled contrast score to the positive class probability and entities ordered in descending rank order.

\subsubsection{Implementation of neural network models}
All preprocessing was done using custom Bash and Python scripts. TransE, RotatE, DistMult, and ComplEx link prediction models were implemented in PyTorch using the DGL-KE package~\cite{dglke} for learning large-scale knowledge graph embeddings. The BERT models were based on HuggingFace BERT implementations using PyTorch. Pre-trained weights for BioBERT (BioBERT-Base v1.1 (+ PubMed 1M))\footnote{\url{https://github.com/naver/biobert-pretrained}}, BioClinicalBERT\footnote{\url{https://huggingface.co/emilyalsentzer/Bio\_ClinicalBERT}}, PubMedBERT\footnote{\url{https://huggingface.co/microsoft/BiomedNLP-PubMedBERT-base-uncased-abstract}} and BlueBERT (BlueBERT-Base, Uncased, PubMed+MIMIC-III)\footnote{\url{https://github.com/ncbi-nlp/bluebert}} came from various sources associated with each paper. We implemented STELP ourselves using a combination of a HuggingFace BERT model and PyTorch. 

\subsection{Discovery patterns}
Discovery patterns are defined as a set of constraints that need to be satisfied for the discovery of new relations between concepts~\cite{hristovski06}. Herein, we used discovery patterns for two purposes. First, we explored an open discovery pattern to identify drugs that can be repurposed for COVID-19. Second, we used the same pattern in closed discovery to propose plausible mechanisms for drugs identified via knowledge graph completion methods described above. Discovery patterns are expressed in terms of predication pairs (or predication chains). In particular, we focused on the following discovery pattern: \\
\\
\texttt{DrugA}-\textsc{inhibits}$|$\textsc{interacts\_with}-\texttt{ConceptB} \textsc{and} \\
\texttt{ConceptB}-\textsc{affects}$|$\textsc{causes}$|$\textsc{predisposes}$|$\textsc{associated\_with}-\texttt{COVIDConcept} 
\textsc{and} \textsc{not} (\texttt{DrugA}-\textsc{treats}-\texttt{COVIDConcept}) 
\\
\\
where \texttt{DrugA} is a drug concept with the semantic type Pharmacologic Substance and \texttt{COVIDConcept} refers to one of the following UMLS concepts (C5203670: \texttt{COVID-19},  C5203676: \texttt{2019 novel coronavirus}, C5203671: \texttt{suspected covid 19}). \texttt{ConceptB} can be any concept, and $|$ indicates logical \textsc{or}. When \texttt{DrugA} is unknown, this corresponds to an open discovery pattern. We used a Neo4j graph database of semantic relations and browser front-end for our exploration.


\subsection{Evaluation}

\subsubsection{Ground truth generation}
We semi-automatically generated a ground truth drug list, similar to the approach in other computational drug repurposing studies for COVID-19 \cite{zeng20}. We downloaded the interventions used in COVID-19 drug trials from clinicaltrials.gov using the following query: 
\url{https://clinicaltrials.gov/ct2/results?cond=COVID-19&term=EXPAND[Term]+COVER[FullMatch]+AREA[InterventionType]+\%22Drug\%22.}

This search yielded a set of 1167 clinical trials. We extracted all the interventions used in these studies and mapped the intervention terms to UMLS CUIs using MetaMap (v2016)~\cite{aronson10} and filtered the resulting concepts by their semantic groups \cite{mccray01}, keeping only those concepts with the semantic group Chemicals \& Drugs. Additionally, we considered the semantic types Therapeutic Procedure and Gene or Genome, which also appeared for some concepts in intervention lists.
We removed the duplicates from the resulting concept list and some general concepts (e.g., \texttt{Therapeutic procedure}, \texttt{Placebo}) as well as incorrect mappings. \textcolor{black}{Drug concepts that only differ in their dosage or mode of administration were grouped together and considered a single element in the ground truth. For example, concepts \texttt{ruxolitinib}, \texttt{ruxolitinib Oral Tablet}, and \texttt{ruxolitinib 5 MG} were clustered together. This pruning and clustering process resulted in a final list of 283 concept clusters.} The automatic evaluation described below was performed against this set.

\subsubsection{Time slicing}
Time slicing is an evaluation technique often used in LBD and link prediction tasks~\cite{henry17}. The idea is to train models on data prior to a specific date and test them on data after that date and evaluate whether links that formed only after the cutoff date can be predicted from the trained model. In this study, we trained our models on semantic relations extracted from publications dated 03/11/2020 or earlier and tested whether they can predict the drugs that have been proposed for COVID-19 since then or have been evaluated in clinical trials. This date was selected as cutoff, as it is the date on which WHO declared COVID-19 a pandemic. It is also a date by which enough biological knowledge about SARS-CoV-2 had accumulated, although COVID-19 therapies were still in their infancy, making it a suitable cutoff for time slicing experiments. 

All five knowledge graph completion models were automatically assessed using an evaluation protocol proposed by Bordes et al.~\cite{transe}. Suppose that $\mathcal{X}$ is a set of triples, $\Theta_E$ be the embeddings of entities $\mathcal{E}$, and $\Theta_R$ be the embeddings of relations $\mathcal{R}$. In the first, corruption step, we go through a set of triples and for each triple $\mathbf{x} = (h, r, t) \in \mathcal{X}$ replace its head and tail with all other entities in $\mathcal{E}$. Each triple is corrupted exactly $2 \vert \mathcal{E} \vert - 1$ times. Formally, the corrupted triple is defined as:
\begin{equation*}
\widetilde{\mathbf{x}} = \bigcup_{h^{\prime} \in \mathcal{E}} (h^{\prime}, r, t) \cup \bigcup_{t^{\prime} \in \mathcal{E}} (h, r, t^{\prime}),
\end{equation*}
where $h^{\prime} \neq h$ and $t^{\prime} \neq t$. We employ the filtered setting protocol not taking into account any corrupted triple that already appears in the knowledge graph. In the second, scoring phase, original and corrupted triples are tested using the constructed classifier $\psi$. Intuition behind this is that the model will assign a higher score to the original triple and a lower score to the corrupted triple. In the third, evaluation phase, the proposed models are assessed using three measures: mean rank (MR), mean reciprocal rank (MRR), and Hits@$k$ measure. MR is an average rank assigned to the true relation, over all relations in a test set:
\begin{equation*}
\mathrm{MR} = \frac{1}{2 \vert \mathcal{T} \vert} \sum_{i = 1}^{\vert \mathcal{T} \vert} \left( \mathsf{rank}_i^h + \mathsf{rank}_i^t \right)
\end{equation*}
where $\mathsf{rank}_i^h$ and $\mathsf{rank}_i^t$ denote the rank position:
\begin{align*}
\mathsf{rank}_i^h &= 1 + \sum_{\widetilde{\mathbf{x}}_i \in \mathcal{C}^h(\mathbf{x}_i) \setminus \mathcal{G}} I \big[ \psi(\mathbf{x}_i) < \psi(\widetilde{\mathbf{x}}_i) \big] \\
\mathsf{rank}_i^t &= 1 + \sum_{\widetilde{\mathbf{x}}_i \in \mathcal{C}^h(\mathbf{x}_i) \setminus \mathcal{G}} I \big[ \psi(\mathbf{x}_i) < \psi(\widetilde{\mathbf{x}}_i) \big],
\end{align*}
where the indicator function $I[P]$ is 1 iff $P$ is true, and 0 otherwise.

MRR is the average inverse rank for all test triples and is formally computed as:
\begin{equation*}
\mathrm{MRR} = \frac{1}{2 \vert \mathcal{T} \vert} \sum_{\mathbf{x}_i \in \mathcal{T}} \frac{1}{\mathsf{rank}_i^h} + \frac{1}{\mathsf{rank}_i^t}
\end{equation*}

Hits@$k$ measures the percentage of relations in which the true triple appears in the top $k$ ranked triples, where $k \in \{1,3,10\}$; formally:
\begin{equation*}
\mathrm{Hits}@k = \frac{100}{2 \vert \mathcal{T} \vert} \sum_{\mathbf{x}_i \in \mathcal{T}} I \Big[ \mathsf{rank}_i^h \leq k \Big] + I \Big[ \mathsf{rank}_i^t \leq k \Big]
\end{equation*}
Our aim was to achieve low MR and high MRR and Hits@$k$. 

\subsubsection{Qualitative evaluation}
In addition, we also performed a qualitative evaluation. One of the authors (MF, MD with a PhD in medical informatics) used Neo4j browser to assess the plausibility of some of the drugs highly ranked by the knowledge completion models, guided by literature search and review, using closed discovery. 
\textcolor{black}{
We also evaluated discovery patterns directly using open discovery. For this purpose, we issued a query for fifty drugs ranked on the number of intermediate \texttt{ConceptB} concepts between the drug and \texttt{COVIDConcept}. Then, MF assessed a subset of the candidate drugs for plausibility.}

\subsubsection{\textcolor{black}{Comparison of candidate drug lists}}
\textcolor{black}{
We compared the drug lists proposed by our methods to each other, as well as to drug lists reported in three prior studies~\cite{zhou20network,zeng20,singh_drug_2020}. For TransE, which performed best, we identified a subset of plausible drugs from its top 150 candidate drug predictions. We used top 50 predictions from other knowledge graph completion methods as well as the top 50 drugs generated using the discovery pattern in open discovery mode.}

\section{Results}
We report the performance of the semantic relation accuracy classifier as well as the knowledge graph completion methods in this section. \textcolor{black}{We also provide a comparison of the drug lists proposed in previous studies and identified by our methods.}

\subsection{Accuracy classifier}

The full table of results for the comparison of various BERT models for the accuracy classifier is included below (Table~\ref{tab:bert}). The chosen model, PubMedBERT, obtained an F$_1$ score of 0.854 (recall = 0.895; precision = 0.816).

\begin{table}[htb!]
\centering
\caption{Results of SemMedDB semantic relation classification using biomedical BERT variants}
\sisetup{table-align-text-post=false,detect-weight=true,detect-inline-weight=math}
\begin{threeparttable}
\begin{tabular}{l*6{S[table-format=1.3,table-column-width=1.7cm]}}
\toprule
& \multicolumn{2}{c}{Vanilla BERT} & {\parbox[c]{17mm}{BioBERT}\tnote{a}} & {\parbox[c]{17mm}{\centering BioClinical\\BERT}} & {\parbox[c]{17mm}{\centering PubMed\\BERT\tnote{b}}} & {BlueBERT\tnote{c}} \\
\midrule
& {Uncased} & {Cased} & {Cased} & {Cased} & {Uncased} & {Uncased}\\
\midrule
\multicolumn{2}{l}{\textit{Validation set}} \\
Rec & 0.815 & 0.767 & 0.861 & 0.822 & \bfseries 0.896 & 0.822 \\
Pre & 0.695 & 0.723 & \bfseries 0.762 & 0.685 & 0.693 & 0.700 \\
F$_1$ & 0.743 & 0.744 & \bfseries 0.808 & 0.748 & 0.781 & 0.756 \\
\midrule
\multicolumn{2}{l}{\textit{Test set}} \\
Rec & 0.815 & 0.782 & 0.842 & 0.832 & \bfseries 0.895 & 0.845 \\
Pre & 0.795 & 0.815 & \bfseries 0.838 & 0.804 & 0.816 & 0.782 \\
F$_1$ & 0.805 & 0.798 & 0.840 & 0.818 & \bfseries 0.854 & 0.812 \\
\bottomrule
\end{tabular}
\begin{tablenotes}[flushleft]
\footnotesize
\note Rec = recall, Pre = precision. Results highlighted in bold are the best for each method.
\item[a] Trained on PubMed 1M
\item[b] Trained on Abstracts + Full text
\item[c] Trained on PubMed + MIMIC
\end{tablenotes}
\end{threeparttable}
\label{tab:bert}
\end{table}

The best model (i.e., PubMedBERT) was then applied to the \num{2558935} predications. Of those, \num{1907717} (74.9\%) were classified as correct predications and retained for use in the training of the downstream models.

This preprocessing yielded \num{115451} unique biomedical concepts and \num{1907717} relations among them. The distribution of these predications are listed in Table~\ref{tab:distribution}.

\begin{table}[htb!]
\sisetup{detect-weight=true,detect-inline-weight=math}
\centering
\caption{Distribution of semantic predications after filtering}
\begin{tabular}{cc|cc}
\toprule
Predicate & Count (\%) & Predicate & Count (\%)\\
\midrule
\textsc{treats} & 518,267 (27.2\%) & \textsc{produces} & 38,602 (2.0\%)\\
\textsc{coexists\_with} & 420,633 (22.1\%) & \textsc{augments} & 37,887 (2.0\%)\\
\textsc{interacts\_with} & 224,809 (11.8\%) & \textsc{prevents} & 25,103 (1.3\%)\\
\textsc{causes} & 205,441 (10.8\%) & \textsc{stimulates} & 24,734 (1.3\%)\\
\textsc{affects} & 192,092 (10.1\%) & \textsc{predisposes} & 18,613 (1.0\%)\\
\textsc{associated\_with} & 106,418 (5.6\%) & \textsc{complicates} & 1,479 (0.1\%)\\
\textsc{inhibits} & 52,518 (2.8\%) & \textsc{manifestation\_of} & 1,156 (0.1\%)\\
\textsc{disrupts} & 39,960 (2.1\%) & &\\
\bottomrule
\end{tabular}
\label{tab:distribution}
\end{table}

\subsection{Knowledge graph completion}

The knowledge graph completion results for all employed models are presented in Table~\ref{tab:linkpred}. For MR a lower score is considered better, for all others a higher score is considered better. The score for each method is the mean value over all triplets in the testing set.

\begin{table}[htb!]
\centering
\caption{Knowledge graph completion results on combined knowledge graph (SemMedDB + CORD-19)}
\begin{threeparttable}
\sisetup{detect-weight=true,detect-inline-weight=math}
\begin{tabular}{lS[table-format=3.3]*4{S[table-format=1.3]}}
\toprule
& {MR} & {MRR} & {Hits@1} & {Hits@3} & {Hits@10}\\
\midrule
TransE &  \bfseries 9.223 & \bfseries 0.525 & \bfseries 0.417 &  \bfseries 0.585 & \bfseries 0.699 \\
DistMult &  11.639 & 0.325 & 0.216 & 0.340 & 0.515 \\
ComplEx &  11.045 & 0.332 & 0.216 & 0.352 & 0.553 \\
RotatE &  10.864 & 0.377 & 0.246 & 0.428 & 0.633 \\
\textcolor{black}{STELP} & \textcolor{black}{22.960} & \textcolor{black}{0.073} & \textcolor{black}{0.000} & \textcolor{black}{0.027} & \textcolor{black}{0.234} \\
\bottomrule
\end{tabular}
\begin{tablenotes}[flushleft]
\footnotesize
\note MR = mean rank, MRR = mean reciprocal rank. Results highlighted in bold are the best for each method.
\end{tablenotes}
\end{threeparttable}
\label{tab:linkpred}
\end{table}

On average, TransE outperforms all counterparts on all performance measures. Optimal TransE configuration was achieved with $k = 400$ hidden dimensions, $L_1$ norm, learning rate $\eta = 0.01$ and regularization coefficient $\lambda = 2 \times 10^{-8}$. Model training was limited to \num{20000} epochs. Relatively small number of relations (15) ensure that all entities and relations can be smoothly embedded into the same vector space. 

\subsection{Embedding representation of knowledge graph}
\begin{figure}
\centering
\includegraphics[width=\textwidth]{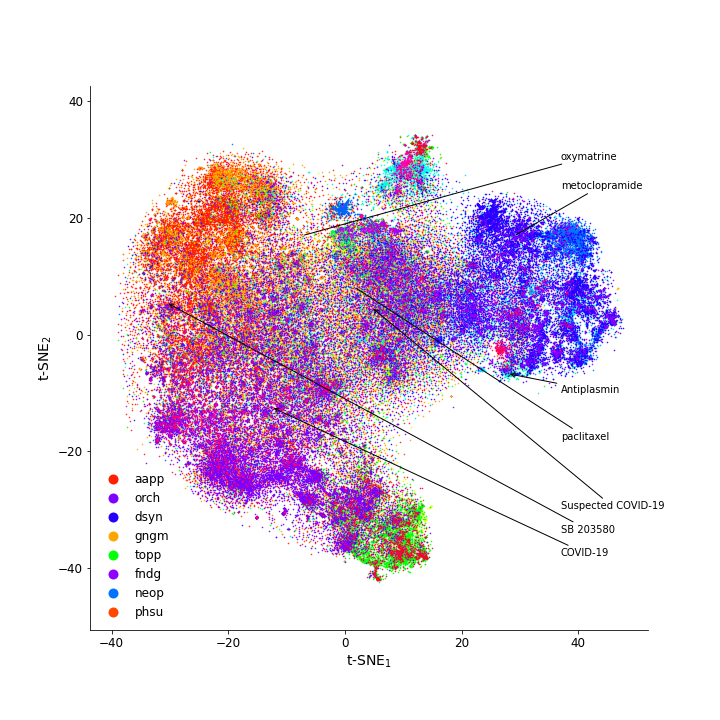}
\caption{Visualization of biomedical concepts learned by t-SNE (t-distributed stochastic neighbor embedding) algorithm and embedded in a two-dimensional space. We highlighted five drugs identified as potential new drugs to treat COVID-19. Color refers to semantic type of a particular concept; note that only the eight most frequent semantic types are presented. aapp: Amino Acid, Peptide, or Protein; dsyn: Disease or Syndrome; fndg: Finding, gngm: Gene or Genome; neop: Neoplastic Process; orch: Organic Chemical; phsu: Pharmacologic Substance; topp: Therapeutic or Preventive Procedure.}
\label{fig:fig4}
\end{figure}

Next, we use t-SNE (t-distributed stochastic neighbor embedding)~\cite{Maaten2008} algorithm to graphically represent embeddings of computed concepts in a two-dimensional space (Figure~\ref{fig:fig4}). t-SNE algorithm enables reduction of high-dimensional data into a low-dimensional space such that similar concepts are presented by nearby points. The plot demonstrates relatively good co-localization of selected concepts, especially for \texttt{Suspected COVID-19} and \texttt{paclitaxel}.

{\subsection{\textcolor{black}{Comparison of proposed drug lists}}
\textcolor{black}{33 drugs (out of top 150) identified by TransE were deemed plausible after manual analysis (Table~\ref{tab:transe}). Comparing this set to the repurposing proposals from three recently published papers \cite{zeng20,zhou20network,singh_drug_2020}, we find that there is one drug in common (estradiol) with the list in Zeng et al.~\cite{zeng20}. On the other hand, Singh et al.~\cite{singh_drug_2020} and Zeng et al.~\cite{zeng20} have eight drugs in common and Zhou et al.~\cite{zhou20network} and Zeng et al.~\cite{zeng20} have three. TransE predictions tended to contain more general drug classes (e.g., anthelmintics, antiplatelet agents), which were not specifically excluded, in contrast to previous methods. On the other hand, it is worth noting that specific drugs in some of these classes have been proposed in other studies. For example, TransE predicted anthelmintics as a candidate, while some of the drugs in this class (e.g., ivermectin, levamisole, nitazoxanide) have been proposed by others and tested in clinical studies. The same can be said about other drug classes, such as mTOR inhibitors and neuraminidase inhibitors. 
}
\begin{table}[htb!]
    \centering
    \caption{\textcolor{black}{33 candidate drugs highly ranked by TransE and deemed plausible in manual analysis.}}
    \begin{threeparttable}
    \begin{tabular}{cc}
    \toprule
        \textcolor{black}{Metoclopramide} & \textcolor{black}{Trilostane}\\
        \textcolor{black}{Oxymatrine}  & \textcolor{black}{Cyproterone Acetate}\\
        \textcolor{black}{\begin{tabular}cMitogen-Activated-\\Protein Kinase Inhibitor\end{tabular}} & \textcolor{black}{\begin{tabular}cNucleoside Reverse-\\Transcriptase Inhibitors\end{tabular}}\\
        \textcolor{black}{Oxophenylarsine} & \textcolor{black}{Methyltrienolone}\\
        \textcolor{black}{5-Alpha reductase inhibitor} & \textcolor{black}{Bosentan}\\
        \textcolor{black}{Folic acid} & \textcolor{black}{Estramustine}\\
        \textcolor{black}{Anthelmintics} & \textcolor{black}{Allicin}\\
        \textcolor{black}{Sildenafil} & \textcolor{black}{Proteasome inhibitors}\\
        \textcolor{black}{Furosemide} & \textcolor{black}{Antiplatelet Agents}\\
        \textcolor{black}{Beclomethasone} & \textcolor{black}{Fibrinolytic Agents}\\
        \textcolor{black}{Cangrelor} & \textcolor{black}{Contraceptive Agents}\\
        \textcolor{black}{Gymnemic acid} & \textcolor{black}{Neuraminidase inhibitor}\\
        \textcolor{black}{Estradiol} & \textcolor{black}{Vitamin D Analogue} \\
        \textcolor{black}{mTOR Inhibitor} & \textcolor{black}{Tyrosine kinase inhibitor}\\
        \textcolor{black}{Clobetasol propionate} & \textcolor{black}{Mometasone furoate}\\
        \textcolor{black}{Carbenoxolone} & \textcolor{black}{Vasopressin Antagonist}\\
        \textcolor{black}{Anti-Retroviral Agents} &\\
    \bottomrule
    \end{tabular}
\end{threeparttable}
\label{tab:transe}
\end{table}

\textcolor{black}{
Comparison of the 33 plausible drugs from TransE with the top 50 predictions from the other knowledge graph completion methods revealed one common drug class between TransE and STELP (5-alpha reductase inhibitors) and five drugs between RotatE and STELP. DistMult and ComplEx did not share any predictions with the other methods. 
}

\textcolor{black}{
Interestingly, using the discovery pattern in open discovery mode, we identified several drugs common with other methods: estradiol with TransE, paclitaxel with RotatE, as well as hydrocortisone and indomethacin with Zeng et al. \cite{zeng20}. Table~\ref{tab:overlaps} lists the overlapping candidate drugs for different methods.}

\begin{table}[htb!]
    \centering
    \begin{threeparttable}
    \caption{\textcolor{black}{Comparison of drug overlap between methods and studies}}
    \begin{tabular}{c|c}
    \toprule
        \textbf{\textcolor{black}{Methods}} & \textbf{\textcolor{black}{Common Drugs}}\\
        \hline
        \begin{tabular}c
            \textcolor{black}{Zeng et al. \cite{zeng20}}\\
            \textcolor{black}{TransE}\\
            \textcolor{black}{Discovery Patterns}\\
        \end{tabular} & \textcolor{black}{Estradiol}\\
        \hline
        \begin{tabular}c
            \textcolor{black}{Zeng et al. \cite{zeng20}}\\
            \textcolor{black}{Singh et al. \cite{singh_drug_2020}}\\
            \textcolor{black}{RotatE}\\
        \end{tabular} & \textcolor{black}{Dexamethasone}\\
        \hline
        \begin{tabular}c
            \textcolor{black}{Zeng et al. \cite{zeng20}}\\
            \textcolor{black}{Discovery Patterns}\\
        \end{tabular} &
        \begin{tabular}c
            \textcolor{black}{Hydrocortisone}\\
            \textcolor{black}{Indomethacin}\\
        \end{tabular}\\
        \hline
        \begin{tabular}c
            \textcolor{black}{Zeng et al. \cite{zeng20}}\\
            \textcolor{black}{STELP}\\
        \end{tabular} & \textcolor{black}{Zidovudine}\\
        \hline
        \begin{tabular}c
            \textcolor{black}{TransE}\\
            \textcolor{black}{STELP}\\
        \end{tabular} & \textcolor{black}{5-alpha Reductase Inhibitors}\\
        \hline
        \begin{tabular}c
            \textcolor{black}{RotatE}\\
            \textcolor{black}{STELP}\\
        \end{tabular} &
        \begin{tabular}c
            \textcolor{black}{Pibrentasvir}\\
            \textcolor{black}{Anti-ILDR2 Monoclonal-}\\
            \textcolor{black}{Antibody BAY 1905254}\\
            \textcolor{black}{Mood Stabilizer}\\
            \textcolor{black}{Opium alkaloids and-}\\
            \textcolor{black}{derivative combination-}\\
            \textcolor{black}{cough suppressants}\\
            \textcolor{black}{Valoctocogene roxaparvovec}\\
        \end{tabular}\\
        \hline
        \begin{tabular}c
            \textcolor{black}{RotatE}\\
            \textcolor{black}{Discovery Patterns}\\
        \end{tabular} & \textcolor{black}{Paclitaxel}
        \label{tab:overlaps}\\
    \tabularnewline
    \bottomrule
    \end{tabular}
    \begin{tablenotes}[flushleft]
    \footnotesize
    \note \textcolor{black}{Drugs are from the top 50 ranked drugs from RotatE, STELP, the 33 drugs from TransE identified by MF as plausible, and the drugs specified in Zeng et al. \cite{zeng20}, Zhou et al. \cite{zhou20network}, and Singh et al. \cite{singh_drug_2020}. We also use top 50 drugs identified using the discovery pattern in open discovery mode.}
    \end{tablenotes}
    \end{threeparttable}
\end{table}

\section{Discussion}
\subsection{Knowledge graph completion models}
\textcolor{black}{Thus far, the following classes of drugs have been used for the management of COVID-19: antivirals, monoclonal antibodies, anti-inflammatory agents, immunomodulators, anticoagulants, and adjuvants \cite{sanders20,wiersinga20}. In addition, several trials have studied antimalarials and antiparasites.}

\textcolor{black}{The knowledge graph completion models predicted drugs in all these classes, although they did not always rank them highly. For example, TransE predicted ribavirin (antiviral), trastuzumab (monoclonal antibody), indomethacin (anti-inflammatory), interferon beta-1b (immunomodulator), heparin (anticoagulant), vitamin D (adjuvant), metronidazole (antiparasite), and artemisone (antimalarial). Dexamethasone, one of the drugs considered most effective for reducing mortality in patients receiving oxygen, was the highest ranking drug from the RotatE model. Results from TransE and RotatE were a mix of individual drugs and drug classes (with little overlap), whereas STELP predictions were largely limited to very specific drugs and also included natural substances such as bioflavonoid quercetin and riboflavin (vitamin B2). While the quantitative evaluation against clinical trial data suggests TransE as the best-performing model, it is worth noting that this only measures how well a method predicts drugs that are currently being trialed. It is difficult to assess the ultimate clinical effectiveness of the proposed drugs, and it is possible that models that do not perform as well quantitatively yield results that prove more promising (as in the case of RotatE and dexamethasone). Despite these issues, qualitative assessment of knowledge graph completion models showed that all methods could identify useful repurposing candidates.}

The results indicate that more complex knowledge graph completion models might not be very efficient in drug repurposing tasks. Due to its relative simplicity, it might be expected that TransE be outperformed by its successors~\cite{rotate,distmult,complex}. However, it showed efficiency in embedding a large-scale complex biomedical knowledge graph, such as the extended SemMedDB used here. On the other hand, differences in performances among DistMult, ComplEx, and RotatE were relatively small. All three models achieved low performance on MRR, Hits@1, and Hits@3 measures, and moderate score on Hits@10. Empirical evidence shows that DistMult and ComplEx usually perform well for high-degree entities, but fails with low-degree entities~\cite{Nguyen2018}. Because we eliminated highly frequent concepts due to their lack of informativeness, it is possible that this is reflected in lower performance of both models.

The context-encoding model, STELP, showed rather poor performance in evaluation. One possibility is that the model was only able to learn high-level groupings for the predicates. This is likely the case as it was observed the model produced much higher scores ($\mathrm{MR}=3.740$, $\mathrm{MRR}=0.867$, $\mathrm{Hits@1}=0.792$, $\mathrm{Hits@10}=0.969$) when evaluating a mix of corrupted triples containing other predicates in addition to \textsc{treats}. Thus, it may be the case that while the model can discriminate between what subjects are feasible for \textsc{treats}-\texttt{COVID-19} versus \textsc{affects}-\texttt{COVID-19} etc., it did not learn more granular features that allow it to differentiate between subjects within the context of \textsc{treats}-\texttt{COVID-19}. However, analysis of the t-SNE embedding and the qualitative evaluation show that the model mostly clustered the ground truth drugs into a couple of large clusters. 

To further compare the drug rankings between TransE and STELP, we performed the Wilcoxon signed-rank test, which shows no correlation between how the two models were ranking novel relations \textcolor{black}{($p = 0.846$)}. Spearman's rank correlation between the novel relation rankings for both models was found to be \textcolor{black}{$-0.004$}, which further supports the results of the Wilcoxon test. Table~\ref{tab:rankStat} and Table~\ref{tab:rankSum} show that there is very little agreement between TransE and STELP, particularly in the top 1000 rankings for each model. It is worth noting that there were 47 items in common in the top 1000 rankings for both models. 

\begin{table}[htb!]
\sisetup{detect-weight=true,detect-inline-weight=math}
\centering
\caption{Statistics for absolute differences of TransE and STELP rankings}
\begin{threeparttable}
\begin{tabular}{cccc} 
\toprule
    & Median & Mean & \begin{tabular}
    [c]{@{}c@{}}Standard\\Deviation
    \end{tabular}  \\
\midrule
Top 1000 TransE Rankings & \textcolor{black}{10789.0} & \textcolor{black}{10567.140} & \textcolor{black}{6128.881}\\
Top 1000 STELP Rankings & \textcolor{black}{10224.0} & \textcolor{black}{10420.0} & \textcolor{black}{6002.522}\\
All Rankings & \textcolor{black}{6342.0}  & \textcolor{black}{7207.910}  & \textcolor{black}{5070.927}\\
\bottomrule
\end{tabular}
\begin{tablenotes}[flushleft]
\footnotesize
\note The values for the first two rows are calculated by taking the top 1000 ranked triples for the specified model, calculating the absolute difference between the rankings from the two models for each of those triples, and calculating the statistics. For example, the triples that TransE ranked as the top 1000 triples were gathered, the absolute differences of rankings between TransE and STELP for those 1000 triples were calculated, and the statistics were calculated from those differences.
\end{tablenotes}
\end{threeparttable}
\label{tab:rankStat}
\end{table}

\begin{table}[htb!]
\sisetup{detect-weight=true,detect-inline-weight=math}
\centering
\begin{threeparttable}
\caption{Summary of absolute differences for TransE and STELP rankings. Semantic types are aapp: Amino Acid, Peptide, or Protein; gngm: Gene or Genome; orch: Organic Chemical; sosy: Sign or Symptom; topp: Therapeutic or Preventive Procedure.}
\begin{tabular}{ccc}
\toprule

\begin{tabular}[c]{@{}c@{}}Max Absolute\\Difference\end{tabular} & Count (\%) & \begin{tabular}[c]{@{}c@{}}Top 3 Most Common\\Semantic Types\end{tabular}\\
\midrule
0 & \textcolor{black}{1 (0.005\%)} & \textcolor{black}{aapp}\\
1 & \textcolor{black}{1 (0.005\%)} & \textcolor{black}{aapp}\\
3  & \textcolor{black}{5 (0.023\%)} & \textcolor{black}{orch, topp, aapp}\\
10 & \textcolor{black}{15 (0.070\%)} & \textcolor{black}{gngm, aapp, orch}\\
100 & \textcolor{black}{189 (0.877\%)} & \textcolor{black}{gngm, aapp, orch}\\
500 & \textcolor{black}{973 (4.516\%)}  & \textcolor{black}{gngm, aapp, orch}\\
1000 & \textcolor{black}{1937 (8.990\%)} & \textcolor{black}{gngm, aapp, orch}\\
\bottomrule
\label{tab:rankSum}
\end{tabular}
\begin{tablenotes}[flushleft]
\footnotesize
\note Count column represents the number of triples where the two models rankings differed by at most the corresponding value in the Max Absolute Difference column. For example, there were 4 triples where both models rankings for those triples differed by at most 3.
\end{tablenotes}
\end{threeparttable}
\end{table}

\subsubsection{Computational efficiency}
The TransE and RotatE are much faster to train than the STELP model (approximately 15 minutes vs. 5 days on our dataset). 
Due to the size of BERT, which lies at the core of the architecture, STELP is a computationally expensive model which makes hyperparameter tuning difficult. This difficulty is compounded on the link prediction task which requires STELP to perform, just for inference, $\mathcal{O}((L/2)^{2}|\mathcal{V}|(1+|\mathcal{E}|))$ steps, where $L$ is the sequence length, $\mathcal{V}$ the number of vertices, and $\mathcal{E}$ the number of edges. As the base BERT model contains 110 million parameters, adding in the scale of the link prediction may make the STELP and similar context-encoding based models infeasible for limited resource settings. TransE and RotatE demonstrate good results at a small fraction of the required computing power and time compared to the STELP model. Due to their reduced required computation time, it can be possible to explore larger graphs than that explored in this work. On the other hand, with adequately large computational resources, it may be possible to optimize STELP hyperparameters and train over multiple random seeds to generate a model that obtains better results than TransE or RotatE, which are limited by their smaller representational capacity.

\subsection{Discovery patterns}
Discovery patterns based on semantic relations provide an intuitive way of exploring potential mechanistic links between biological phenomena. Neo4j and its query language, Cypher, are powerful tools that complement semantic relations nicely in quickly pinpointing promising research directions, although massive graphs present some challenges for effective query and retrieval. In addition, a human expert is needed to sort out the noise in semantic relations (some of it obvious) due to NLP errors. However, given that predictions made by the knowledge completion models above are largely opaque, a human-in-the-loop discovery browsing approach based on patterns \cite{wilkowski11,cairelli13} remains an effective alternative to these more complex approaches, and also complements them by providing potential explanations. \textcolor{black}{Given the size of the graph and time constraints, we limited ourselves to a single discovery pattern in this study and were able to both identify promising drugs (\emph{open discovery}) and generate potential explanations for drugs predicted by the knowledge graph completion methods (\emph{closed discovery}).}


\textcolor{black}{Using the open discovery pattern approach, we identified five promising drugs that were ranked highly and were not, to our knowledge, discussed in the literature (paclitaxel, SB 203580, alpha 2-antiplasmin, pyrrolidine dithiocarbamate, and butylated hydroxytoluene). The same approach also ranked highly some drugs and substances  evaluated in clinical trials (e.g., quercetin, melatonin, vitamin D, estradiol, and simvastatin). We discuss three that have not been proposed for COVID-19 in more detail below (paclitaxel, SB 203580, and alpha 2-antiplasmin). Notably, paclitaxel as well as quercetin, melatonin, vitamin D, estradiol, and simvastatin were predicted by the knowledge graph completion models.}
Figure~\ref{fig5} shows a network resulting from the aforementioned discovery pattern generated by Neo4j browser.

\textcolor{black}{We also used discovery patterns to generate mechanistic explanations for two other drugs ranked highly by TransE and deemed plausible. These drugs are metoclopramide and oxymatrine, which are also discussed in more detail below.}

\begin{figure}
\centering
\includegraphics[width=\textwidth]{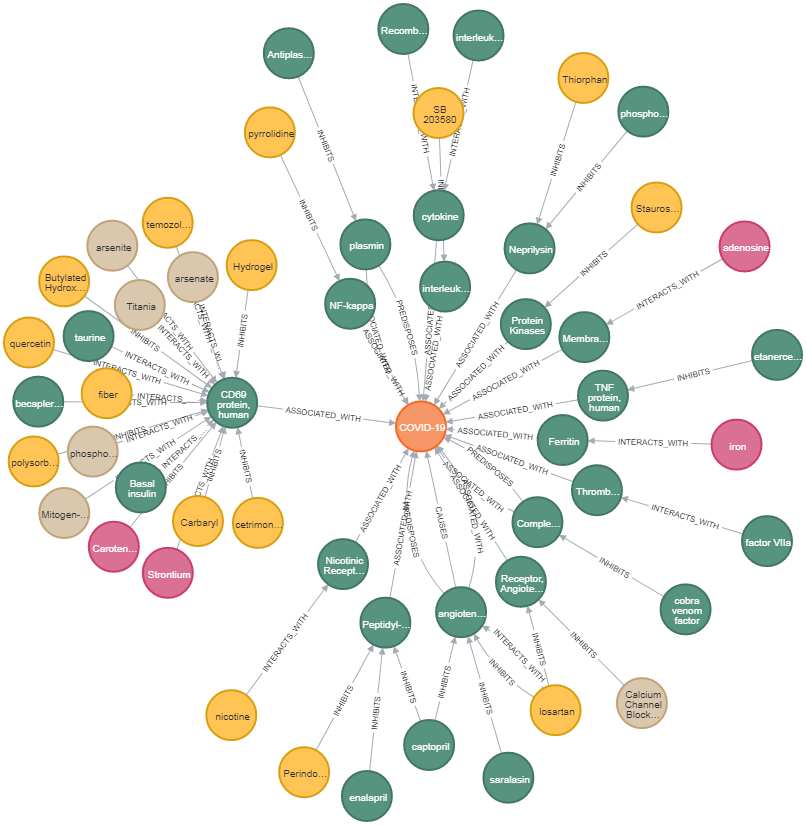}
\caption{Drug repurposing for COVID-19 with the open discovery pattern \texttt{DrugA}-\textsc{inhibits}$|$\textsc{interacts\_with}-\texttt{ConceptB} \textsc{and} \texttt{ConceptB}-\textsc{affects}$|$\textsc{associated\_with}$|$\textsc{causes}$|$\textsc{predisposes}-\texttt{COVID-19}. The directionality is from the periphery (the predicted drugs) through the intermediate concepts to COVID-19 in the center.}
\label{fig5}
\end{figure}

\subsubsection{Paclitaxel}
Paclitaxel is used to treat several cancer types, including ovarian cancer, breast cancer, lung cancer, cervical cancer, and pancreatic cancer. It stabilizes the microtubule polymer and protects it from disassembly, rendering chromosomes unable to achieve a metaphase spindle configuration. This blocks the progression of mitosis and prolonged activation of the mitotic checkpoint triggers apoptosis or reversion to the G0-phase of the cell cycle without cell division \cite{weaver14}. The following patterns support the paclitaxel discovery:
\begin{enumerate}
\item \texttt{paclitaxel}-\textsc{inhibits}-\texttt{interleukin-6}-\textsc{causes}-\texttt{COVID-19}
\item \texttt{paclitaxel}-\textsc{inhibits}-\texttt{NF-kappa B}-\textsc{associated\_with}-\texttt{COVID-19}
\item \texttt{paclitaxel}-\textsc{inhibits}-\texttt{interleukin-1, beta}-\textsc{associated\_with}-\texttt{COVID-19}
\item \texttt{paclitaxel}-\textsc{inhibits}-\texttt{Granulocyte Colony-Stimulating Factor}-\\\textsc{associated\_with}-\texttt{COVID-19}
\item \texttt{paclitaxel}-\textsc{inhibits}-\texttt{interleukin-10}-\textsc{predisposes}-\texttt{COVID-19}
\item \texttt{paclitaxel}-\textsc{inhibits}-\texttt{interleukin-8}-\textsc{predisposes}-\texttt{COVID-19}
\item \texttt{paclitaxel}-\textsc{inhibits}-\texttt{Thromboplastin}-\textsc{associated\_with}-\texttt{COVID-19}
\item \texttt{paclitaxel}-\textsc{interacts\_with}-\texttt{TLR4}-\textsc{causes}-\texttt{COVID-19}
\end{enumerate}

The first six patterns support a role for paclitaxel in alleviating the cytokine storm of COVID-19, triggered by dysfunctional immune response and mediating widespread lung inflammation. 
Paclitaxel may plausibly help as an  immunosuppressive therapy to immunomediated damage in COVID-19 \cite{tay20}.
Thromboplastin (pattern 7) is a complex enzyme found in brain, lung, and other tissues and especially in blood platelets and functions in the conversion of prothrombin to thrombin in the clotting of blood and may be elevated in patients with COVID-19. As pulmonary microvascular thrombosis plays an important role in progressive lung failure in COVID-19 patients, paclitaxel may reduce the state of hypercoagulability by acting as an inhibitor of thromboplastin \cite{miesbach20}.
The final pattern involves the interaction of paclitaxel with TLR4. Paclitaxel is known to have high affinity for TLR4 receptors. SARS-CoV-2 Spike protein binds with human innate immune receptors, mainly TLR4, increasing secretion of IL‐6 and TNF‐$\alpha$ and neuroimmune response. This suggests that paclitaxel may dislocate SARS-CoV-2 Spike proteins \cite{ran15,brandao20}.

\textcolor{black}{We note that paclitaxel, as a chemotherapy drug, is associated with adverse effects, some serious, such as neutropenia, leukopenia, alopecia, arthralgia, myalgia, and peripheral neuropathy \cite{paclitaxel_dailymed}}.

\subsubsection{SB 203580}
SB 203580 is a specific inhibitor of p38$\alpha$, which suppresses downstream activation of MAPKAP kinase-2, involved in many cellular processes including stress and inflammatory responses and cell proliferation. The following patterns support the SB 203580 discovery:
\begin{enumerate}
\item \texttt{SB 203580}-\textsc{inhibits}-\texttt{interleukin-6} -\textsc{causes}-\texttt{COVID-19}
\item \texttt{SB 203580}-\textsc{inhibits}-\texttt{TNF protein, human}-\textsc{associated\_with}-\texttt{COVID-19}
\item \texttt{SB 203580}-\textsc{inhibits}-\texttt{interleukin-1, beta}-\textsc{associated\_with}-\texttt{COVID-19}
\item \texttt{SB 203580}-\textsc{inhibits}-\texttt{interleukin-8}-\textsc{predisposes}-\texttt{COVID-19}
\item \texttt{SB 203580}-\textsc{inhibits}-\texttt{NF-kappa B}-\textsc{associated\_with}-\texttt{COVID-19}
\item \texttt{SB 203580}-\textsc{inhibits}-\texttt{Interleukin-1}-\textsc{causes}-\texttt{COVID-19}
\item \texttt{SB 203580}-\textsc{inhibits}-\texttt{Granulocyte-Macrophage Colony-Stimulating Factor} -\textsc{associated\_with}-\texttt{COVID-19}
\item \texttt{SB 203580}-\textsc{inhibits}-\texttt{Interleukin-17}-\textsc{associated\_with}-\texttt{COVID-19}
\item \texttt{SB 203580}-\textsc{inhibits}-\texttt{Macrophage Colony-Stimulating Factor}-\\
\textsc{associated\_with}-\texttt{COVID-19}
\end{enumerate}

Similarly to paclixatel, all patterns involving SB 203580 point to a potential inhibition of the hyperinflammatory response in COVID-19. According to Gaestel \cite{gaestel13}, ``the role of the protein kinases p38$\alpha$ in inflammation and innate immunity was found when the compound SB 203580 suppressed tumor necrosis factor (TNF) production in monocytes, and this resulted in inhibition of septic (infammatory) shock.''

\subsubsection{Alpha 2-antiplasmin}
Alpha 2-antiplasmin is a serine protease inhibitor responsible for inactivating plasmin. Elevated plasmin is a common risk factor for COVID-19 susceptibility, especially in patients with comorbidities such as hypertension, diabetes, and coronary heart disease \cite{ji20plasmin}. 
The following patterns support the alpha 2-antiplasmin discovery:
\begin{enumerate}
    \item \texttt{Alpha 2-antiplasmin}-\textsc{inhibits}-\texttt{plasmin}-\textsc{predisposes}-\texttt{COVID-19}
    \item \texttt{Alpha 2-antiplasmin}-\textsc{inhibits}-\texttt{fibrinogen}-\textsc{associated\_with}-\texttt{COVID-19}
    \item \texttt{Alpha 2-antiplasmin}-\textsc{interacts\_with}-\texttt{IgY}-\textsc{associated\_with}-\texttt{COVID-19}
\end{enumerate}

More specifically, plasmin may cleave a newly inserted furin site in the S protein of SARS-CoV-2, which increases its infectivity and virulence in COVID-19. In addition, fibrinogen levels are higher in COVID-19 patients and may contribute to hypercoagulability \cite{ji20plasmin}. By inhibiting plasmin and fibrinogen (first two patterns), Alpha 2-antiplasmin may confer protection to COVID-19. In addition, pattern 3 suggests a mechanism of protection via immunoglobulin Y (IgY). In the immunology field, IgY against acute respiratory tract infection has been developed for more than 20 years. Several IgY applications have been effectively confirmed in both human and animal health. IgY antibodies extracted from chicken eggs have been used in bacterial and viral infection therapy. IgY production has been proposed as immunization as an adjuvant therapy in viral respiratory infection caused by COVID-19 infection \cite{constantin20}. Chicken immunized with Alpha 2-antiplasmin and the peptide-specific antibody (IgY) was isolated from the egg yolks of hens that could be used as potential protections for COVID-19 patients \cite{lee97purification}.

\subsubsection{Metoclopramide}

\textcolor{black}{
Metoclopramide is used to relieve symptoms such as nausea, vomiting, and heartburn, caused by gastroesophageal reflux disease or diabetic gastroparesis. Metoclopramide is, mostly, a dopamine D2 antagonist but acts on many other neurotransmitters and proteins. Using our discovery pattern, we identified two pathways through which metoclopramide may protect against COVID-19.}
\textcolor{black}{
\begin{enumerate}
    \item \texttt{metoclopramide}-\textsc{interacts\_with}-\texttt{cholinergic system}-\\\textsc{associated\_with}-\texttt{COVID-19}
    \item \texttt{metoclopramide}-\textsc{inhibits}-\texttt{TNF protein, human}-\\\textsc{associated\_with}-\texttt{COVID-19}
\end{enumerate}
}

\textcolor{black}{
The first pattern suggests a cholinergic pathway for the protective effect of metoclopramide. The first relation of this pattern is extracted from a study which suggests that metoclopramide activates the sympathetic nervous system by mediating the central cholinergic system in humans \cite{takeuchi93}.  The second piece of the evidence is based on a paper that explains how a cholinergic anti-inflammatory pathway acting through acetylcholine receptors can inhibit the production of pro-inflammatory cytokines \cite{tizabi20}. Therefore, by activating the cholinergic pathway, metaclopropamide may prevent the inflammatory cytokine storm associated with COVID-19.}
}

\textcolor{black}{
The second potential link is via tumor necrosis factor-$\alpha$ (TNF-$\alpha$), a cytokine used by the immune system for cell signaling. The inhibitory effect of metoclopramide on TNF-$\alpha$ is suggested by a study on anti-inflammatory properties of benzamides, a class of drugs to which metoclopramide belongs (``Our data have shown that metoclopramide \dots gave dose dependent inhibition of TNF$\alpha$'' \cite{pero99}). The second piece of the link comes from a paper that studies the role of TNF-$\alpha$ as a key driver of inflammatory macrophage response in severe COVID-19 and proposes anti-cytokine (especially, anti-TNF) treatment for COVID-19 \cite{zhang20ifn}.
}

\subsubsection{Oxymatrine}
\textcolor{black}{
Oxymatrine is a quinazine alkaloid with organ- and tissue-protective effects, primarily related to its anti-inflammatory, anti-oxidative stress, anti- or pro-apoptotic, anti-fibrotic, metabolism-regulating, and anti-nociceptive functions \cite{lan20}. In addition, a variety of signal pathways, cells, and molecules are influenced by oxymatrine.
}

\textcolor{black}{
The following patterns support the repurposing of oxymatrine:
\begin{enumerate}
    \item \texttt{oxymatrine}-\textsc{inhibits}-\texttt{interleukin-6}-\\\textsc{predisposes}-\texttt{COVID-19}
    \item \texttt{oxymatrine}-\textsc{inhibits}-\texttt{TNF protein, human}-\textsc{associated\_with}-\texttt{COVID-19}
    \item \texttt{oxymatrine}-\textsc{inhibits}-\texttt{interleukin-1, beta}-\\\textsc{associated\_with}-\texttt{COVID-19}
    \item \texttt{oxymatrine}-\textsc{inhibits}-\texttt{NF-kappa B}-\\\textsc{associated\_with}-\texttt{COVID-19}
    \item \texttt{oxymatrine}-\textsc{interacts\_with}-\texttt{interleukin-10}-\textsc{predisposes}-\texttt{COVID-19}
    \item \texttt{oxymatrine}-\textsc{inhibits}-\texttt{TLR4 gene}-\\\textsc{associated\_with}-\texttt{COVID-19}
\end{enumerate}
}

\textcolor{black}{%
The first five patterns illustrate the effect of oxymatrine on proinflammatory cytokine and chemokine production induced by SARS-CoV-2. The first piece of the evidence is often an inhibitory relationship, as stated in Huang et al.~\cite{huang12b}: ``Oxymatrine at 120mg/kg significantly suppressed gene expressions of TLR-4 and NF-$\kappa$B, decreased levels of TNF-$\alpha$, interleukin-1beta and interleukin-6''. The relationship between cytokine response and COVID-19 is well-established, for example as stated in Chi et al.~\cite{chi20serum}: ``IL-6, IL-7,  IL-10, \dots were found to be associated with the severity of COVID-19''.  Furthermore, the authors of the latter article propose that immunomodulatory treatment to regulate the cytokine responses could be an effective therapeutic strategy for SARS-CoV-2 infection. Oxymatrine could be one such candidate.}

\textcolor{black}{%
The relevance of TLR4 (Toll-like receptor 4) (pattern 6) for COVID-19, on the other hand, can be gleaned from Choudhury et al. \cite{choudhury20}, which states that ``TLR4 may have a crucial role in the virus-induced inflammatory consequences associated with COVID-19.'' The authors further make the point that TLR4 antagonists (such as oxymatrine) could pave the way for COVID-19 treatment.}


\subsection{Error analysis}
\textcolor{black}{
As error analysis, we manually examined the top 150 predictions by the best-performing model, TransE, for plausibility. 99 of these were deemed implausible, as they were drug classes that were considered too general. Examples of such classes include \texttt{C0003205:} \texttt{Anti-Infective Agents}, \texttt{C0003367: Antilipemic Agents}, and \texttt{C0010858: Cytostatic Agents}. As noted above, some members of these classes may indeed be plausible; however, the classes themselves were considered errors. A more systematic approach to exclude drug classes (e.g., by using MeSH concept hierarchy) could help reduce such errors. A more fine-grained evaluation could also consider such cases as partially correct, although this is unlikely to be useful for drug repurposing. 
}

\textcolor{black}{
The other 18 candidates in the list that were deemed implausible are those that were classified as pharmacologic substances in UMLS, but were not drugs. These include \texttt{C0279328:} \texttt{hyperbaric oxygen}, \texttt{C1618233}: \texttt{husk} and \texttt{C1443923: Oral rehydration}, among others. It may be possible to leverage drug knowledge resources, such as DrugBank, to exclude such concepts and reduce errors.}

\subsection{Limitations and future work}
Our approach relies on accuracy of the predications extracted by SemRep. SemRep precision is about 0.70 and its recall around 0.42~\cite{kilicoglu20semrep}. While  the accuracy classifier helped us improve the accuracy of the predications used, the remaining errors were still significant, impacting the knowledge graph completion task. 

In addition, despite aggressive filtering, the graph formed by the relations in extended SemMedDB is very large, making it difficult to apply computationally intensive models like STELP. In this study, we examined a sub-graph which, inevitably, results in a loss of information available to knowledge graph completion techniques. While we were still able to apply modeling techniques to a fairly large sub-graph focusing on drug repurposing, there exists a larger, complementary sub-graph that may provide further drug candidates.

As noted above, the TransE model benefited from hyperparameter tuning using a grid search method to find an optimal configuration. Similarly, STELP would likely benefit from a similar tuning to find an optimal configuration. For example, a single linear layer was used on the pooled output from the BioBERT model to produce the logits when increasing the representational capacity of the linear layer, by depth or width, might allow for STELP to develop a richer model of the underlying space formed by the BioBERT contextualized embeddings. 

Our methods were limited to knowledge from the literature. Other types of biological data (e.g., protein-protein interactions, drug-target interactions, gene/protein sequences, pharmacogenomic and pharmacokinetic data) are likely to benefit identification of drug candidates, as shown to some extent by other studies \cite{zhou20ai}, as well as our prior work \cite{hristovski10}. However, the computational resources needed for training models based on such massive data can be prohibitive. TransE and similar methods seem more promising in that respect.

Lastly, with our \emph{in silico} approach, we can of course only propose drug candidates for repurposing. To evaluate whether these drugs could indeed act as effective treatments for COVID-19, wet lab experiments and clinical studies are needed. However, the fact that we were able to identify drugs known to have some benefit for COVID-19 (e.g., dexamethasone) via purely computational methods that rely only on automatically extracted literature knowledge is encouraging. \textcolor{black}{Moreover, the use of discovery patterns to explain why a particular drug or substance could be beneficial in prioritizing the most promising candidates for such studies.}

\section{Conclusion}
In this study, we proposed an approach that combines literature-based discovery and knowledge graph completion for COVID-19 drug repurposing. Unlike similar efforts that largely focused on COVID-19-specific knowledge, we incorporated knowledge from a wider range of biomedical literature. We used state-of-the-art knowledge graph completion models as well as simple but effective discovery patterns to identify candidate drugs. We also demonstrated the use of these patterns for generating plausible mechanistic explanations, showing the complementary nature of both methods.

The approach proposed here is not specific to COVID-19 and can be used to repurpose drugs for other diseases. It can also be generalized to answer other clinical questions, such as discovering drug-drug interactions or identifying drug adverse effects.

As COVID-19 pandemic continues its spread and disruption around the globe, we are reminded how the spread of infectious diseases is increasingly common and future pandemics ever more likely. Innovative computational methods leveraging existing biomedical knowledge and infrastructure could help us plan for, respond to, and mitigate the effects of such global health crises. Drug repurposing is a key piece of this response, and our approach provides an efficient computational method to facilitate this goal. 

\section*{Acknowledgments}
We thank Fran\c{c}ois-Michel Lang, Leif Neve, and Jim Mork for their assistance with processing the CORD-19 dataset with SemRep and providing updates to SemMedDB. We acknowledge Tom Rindflesch for his encouragement with the project.

\section*{Funding}
RZ and DS were supported by the U.S. National Institutes of Health’s National Center for Complementary and Integrative Health (Grant No. R01AT009457). DH was supported by Slovenian Research Agency (Grant No. J5-1780, J5-2552, and P3-0154). AK was supported by the Slovenian Research Agency (Grant No. P3-0154, Z5-9352, and J5-2552).

\bibliography{jbi}

\end{document}